\documentclass{article}
\usepackage{ijcai23}

\usepackage{times}
\usepackage{soul}
\usepackage{url}
\usepackage[hidelinks]{hyperref}
\usepackage[utf8]{inputenc}
\usepackage[small]{caption}
\usepackage{graphicx}
\usepackage{amsmath}
\usepackage{amsthm}
\usepackage{amssymb}
\usepackage{booktabs}
\usepackage{algorithm}
\usepackage{algorithmic}
\usepackage[switch]{lineno}

\newtheorem{theorem}{Theorem}

\newtheorem{definition}{Definition}
\newtheorem{lemma}{Lemma}

\title{Cardinality-Minimal Explanations for Monotonic Neural Networks}

\author{
Ouns El Harzli$^1$
\and
Bernardo Cuenca Grau$^1$\and
Ian Horrocks$^1$
\affiliations
$^1$Department of Computer Science, University of Oxford\\
\emails
\{ouns.elharzli, bernardo.cuenca.grau, ian.horrocks\}@cs.ox.ac.uk
}

\begin{document}

\maketitle

\begin{abstract}
    In recent years, there has been increasing interest in explanation methods for neural model predictions that offer precise formal guarantees. These include
    abductive (respectively, contrastive) methods, which aim to compute  minimal subsets
    of input features that are sufficient for a given prediction to hold (respectively, to change
    a given prediction). The corresponding decision problems are, however, known to be intractable. 
    In this paper, we investigate whether  tractability can be regained by focusing on neural models implementing a monotonic function. Although the relevant decision problems remain intractable, we can show that they become solvable in polynomial time by means of greedy algorithms if we additionally assume that the activation functions are continuous everywhere and differentiable almost everywhere. Our experiments suggest favourable performance of our algorithms.


   \end{abstract}

\section{Introduction}

Deep Neural Networks have experienced
unprecedented success in areas such as image analysis, NLP, speech recognition, and data science, with systems  outperforming humans in a wide range of tasks  \cite{deeplearning1,deep2,lecun2015deep,schmidhuber2015deep,deeplearning3}. 
As the use of neural models
becomes  widespread, however, task performance is no longer the only driver of system design, and criteria such as safety, fairness, 
and robustness have gained prominence in recent years \cite{aiethics}. 
Improving model interpretability is an important step towards fulfilling these criteria: if models can explain their predictions, it becomes easier to ensure that they are safe, fair and robust. 
This is, however, notoriously challenging,  
as neural models are `black boxes' where predictions rely on complex numeric calculations. 

A wealth of explanation methods have been proposed in recent years:   \emph{attribution-based methods}
assign a score to input features quantifying their contribution to the prediction relative to a baseline  \cite{axiomaticattrib,manyshapsundararajan20b,attribunderstandingancona2018};  \emph{example-based methods}
explain predictions by retrieving training examples 
that are most similar to the input \cite{influencefunc,prototypes}; and  \emph{perturbation-based methods} generate small corrections to the input causing the output to change \cite{minimaldistance,counterfactualvisual,gnncounterfactual,robustgnncounterfactual}. 
These methods, however, have been criticised for their lack of formal guarantees \cite{DBLP:conf/pods/Darwiche20,DBLP:conf/nips/BlancLT21,DBLP:journals/corr/abs-2211-00541}, which handicaps their applicability to high-risk or safety-critical scenarios. 

As a result, there is increasing interest in
explanation methods
providing rigorous formal guarantees \cite{DBLP:conf/pods/Darwiche20,DBLP:journals/corr/abs-2211-00541,cucala2022explainable,DBLP:conf/aaai/IgnatievNM19,DBLP:conf/ijcai/ShihCD18}. 
\emph{Rule-based methods} generate explanations  in the form of logic rules which
are sufficient to derive a given prediction \cite{cucala2022explainable,contrastivexplain,DBLP:conf/kr/CucalaGM2}.
\emph{Abductive methods} \cite{DBLP:conf/aaai/IgnatievNM19,DBLP:conf/ijcai/ShihCD18,DBLP:conf/nips/BarceloM0S20} 
aim to compute `\emph{Why?}' explanations---minimal subsets of input features that are sufficient for deriving the prediction; dually, \emph{contrastive methods}
compute `\emph{Why Not?}' explanations---minimal subsets of input features so that some
change in their value yields a change in the model's prediction. The formal guarantees provided by these methods are given by both the
soundness requirement (i.e., the explanation is guaranteed to preserve or change the prediction) and the minimality requirement, where minimality can be understood in terms of set inclusion (\emph{subset-minimal explanations}) or number of elements (\emph{cardinality-minimal explanations}); the latter leads to smaller explanations in general since every cardinality-minimal explanation is also subset-minimal but not vice-versa. Furthermore, the size of cardinality-minimal explanations is also related to measures of robustness of neural predictions proposed in the literature \cite{DBLP:conf/kr/ShiSDC20}.

Abductive and constrastive
explanations can be formalised 
as \emph{explainability queries}---Boolean
questions that can be posed to
a model and an input feature vector \cite{DBLP:conf/nips/BarceloM0S20}---,
and the computational complexity of the corresponding decision problem for different types of models can be rigorously 
studied \cite{DBLP:journals/jair/WaldchenMHK21,DBLP:conf/nips/BarceloM0S20}. 
In this paper, we focus on two explainability queries in the context of neural models \cite{DBLP:conf/nips/BarceloM0S20}. The \emph{Minimum Change Required} (MCR) query can be used to compute minimal contrastive explanations: given a model, an input vector $\mathbf{x}$, and $k \in \mathbb{N}$, the query is true if there exists an input vector $\mathbf{y}$ differing from $\mathbf{x}$ in at most $k$ components and which yields a different prediction. Dually, the \emph{Minimum Sufficient Reason} (MSR) query can be used to compute minimal abductive explanations, also called minimal prime-implicants  \cite{DBLP:conf/ijcai/ShihCD18}; given
a model, an input vector $\mathbf{x}$, and $k \in \mathbb{N}$, the query is true if there is a subset 
$\mathbf{y}$ of the components of $\mathbf{x}$
of size at most $k$ that suffices to obtain the current prediction (i.e., such that the prediction remains the same  regardless of the values assigned to the  components not in $\mathbf{y}$).

Unfortunately, MCR and MSR are NP-complete and $\Sigma_2^p$-complete respectively, already for neural
models with ReLU activations implementing a Boolean function \cite{DBLP:conf/nips/BarceloM0S20}.
Although different techniques have been proposed to cope with intractability \cite{DBLP:conf/kr/ShiSDC20,DBLP:conf/aaai/IgnatievIS022,DBLP:conf/ijcai/Izza021},  
these complexity results may still constitute a challenge in practical scenarios.

A way to recover tractability is to 
focus on neural models implementing  functions satisfying additional properties, such as \emph{monotonicity} \cite{DBLP:conf/icml/0001GCIN21,DBLP:journals/ijon/CanoGKWG19,cucala2022explainable}; although 
 this restricts the model's
expressive power, the monotonicity assumption remains appropriate for a wide range of learning tasks. Furthermore, it was shown that 
subset-minimal abductive explanations can be computed in polynomial time if the model implements a monotonic real-valued function \cite{DBLP:conf/icml/0001GCIN21}.

In this paper, we study cardinality-minimal abductive and
contrastive explanations for monotonic neural  architectures.
We first show that MCR and MSR
remain intractable for 
fully-connected neural networks implementing monotonic Boolean functions. Thus, 
although subset-minimal abductive
and contrastive explanations can be computed in polynomial time  
\cite{DBLP:conf/icml/0001GCIN21}, cardinality-minimal abductive or contrastive explanations cannot be efficiently computed under standard complexity assumptions.

Our hardness proofs, however, rely on neural models 
equipped with the step activation function.
We then focus our attention on monotonic neural networks where the non-linear activations are continuous everywhere and differentiable almost everywhere  (as is the case with most practical activations such as ReLU and sigmoid). We show that, in this setting, both
cardinality-minimal contrastive explanations (and hence the MCR query) and abductive explanations (thus the MSR query) can be computed in polynomial time by means of a greedy algorithm. Our tractability results not only apply to models implementing Boolean functions, but also to  more general settings involving real-valued functions.
To show correctness of our greedy algorithms, we exploit 
the theoretical properties of the \emph{integrated gradients} method \cite{axiomaticattrib}, thus establishing a connection between 
the theory of attribution methods developed by the ML community and the theory of abductive and contrastive explanations developed by the KR community. We note, however, that our algorithms do not rely on the application of attribution methods, but only on the
ability to apply the model as a black box.



We conducted experiments on two partially monotonic datasets commonly used as benchmarks for designing monotonic and partially-monotonic models \cite{certifiedmonotonicneurips2020}: Blog Feedback Regression \cite{blogreg}, a regression dataset with 276 features 
and Loan Defaulter\footnote{https://www.kaggle.com/datasets/wordsforthewise/lending-club}, a classification dataset with 28 features. 
We trained monotonic FCNs to reach acceptable performance and we computed both contrastive and abductive explanations. The experiments showed that contrastive explanations are typically of small cardinality, whereas abductive explanations are typically larger, which is expected for a partially monotonic dataset. In both cases, the explanations could be efficiently computed.

\section{Preliminaries and Background}

In this section, we fix the notation used in the remainder of the paper and define the basic neural models that we consider. We also introduce attribution-based methods as well as the MCR and MSR explainability queries underpinning the theoretical analysis of contrastive and abductive explanations. 

\paragraph{Notation.} We let bold-face lowercase letters denote real-valued vectors.  Given vector $\mathbf{x}$, we use $x_i$ to denote
its $i$-th component. 
Given  $\mathbf{x}, \mathbf{x}' \in \mathbb{R}^n$ and a subset $S \subseteq \{1,...,n\}$ of their components, we denote
with $\mathbf{x}^{S|\mathbf{x}'}$ the vector obtained from $\mathbf{x}$ by setting each component $x_i$ with $i \in S$ to $x_i'$.
The complement $\overline{\mathbf{x}}$ of a
a Boolean vector $\mathbf{x} \in \left\{0,1\right\}^n$
is obtained from $\mathbf{x}$ by replacing each $0$ with $1$ and vice-versa. We denote with $\mathbf{0}_m$ the $m$-dimensional column null vector.
We use bold-face capital letters for matrices and denote the $(i,j)$ component of a matrix $\mathbf{M}$ as
$M_{i,j}$. 
Given function $f: \mathbb{R}^n \mapsto \mathbb{R}$, we denote with 
$(\nabla f)_i$ the $i$th component of its
gradient.

\paragraph{Fully-Connected Networks.}
A fully-connected neural network (FCN) with $L \geq 1$ layers and input dimension $n$ is a tuple 
$\mathcal{N} = \langle \{\mathbf{W}^{\ell}\}_{1 \leq \ell \leq L}, \{\mathbf{b}^{\ell}\}_{1 \leq \ell \leq L}, \{\sigma^{\ell}\}_{1 \leq \ell \leq L\}}, \mathrm{cls}  \rangle$.
For each layer $\ell \in \{1, \ldots, L\}$, the integer $d_{\ell} \in \mathbb{N}$ is the \emph{width} of layer $\ell$ and we require $d_{L} = 1$ and define $d_0 = n$; matrix  $\mathbf{W}^{\ell} \in \mathbb{R}^{d_{\ell} \times d_{\ell-1}}$ is a \emph{weight matrix}; vector $\mathbf{b}^{\ell} \in \mathbb{R}^{d_{\ell}}$ is a \emph{bias vector}; $\sigma^{\ell} : \mathbb{R} \mapsto \mathbb{R}$ is a polytime-computable  
\emph{activation function} applied component-wise to vectors; and $\mathrm{cls} : \mathbb{R} \mapsto \{0,1\}$ a polytime-computable and monotonic \emph{classification function}.
The domain $\Delta \subseteq \mathbb{R}^{n}$ of $\mathcal{N}$ specifies the
set of input feature vectors to which the network is applicable.
The application of $\mathcal{N}$ to $\mathbf{x} \in \Delta$ generates a sequence
$\mathbf{x}^1, \ldots, \mathbf{x}^L$ of vectors defined as 
$\mathbf{x}^{\ell} = \sigma^{\ell}(\mathbf{h}^{\ell})$, where $\mathbf{x}^0 = \mathbf{x}$ and
$\mathbf{h}^{\ell}  =   \mathbf{W}^{\ell} \cdot \mathbf{x}^{\ell-1} + \mathbf{b}^{\ell}$.
The result $\mathcal{N}(\mathbf{x})$ of applying $\mathcal{N}$ to  $\mathbf{x}$ is the scalar $\mathrm{cls}(\mathbf{x}^L$).
Thus, the neural network realises a function $\mathcal{N}: \Delta \mapsto \{0,1\}$. We denote with $\Tilde{\mathcal{N}} (\mathbf{x}) := \mathbf{x}^L$ the output of the last layer (before classification). The monotonicity of function $\mathrm{cls}$ implies that there exists a prediction threshold $t := \mathrm{inf}_{z \in \mathbb{R}} \; (\mathrm{cls} (z) = 1)$, such that $\Tilde{\mathcal{N}} (\mathbf{x}) > t$ implies $\mathcal{N} (\mathbf{x}) =1$ and $\Tilde{\mathcal{N}} (\mathbf{x}) < t$ implies $\mathcal{N} (\mathbf{x}) =0$.

When $\sigma^{\ell}$ is the rectified linear unit (ReLU) for each $\ell \in \{1, \ldots, L\}$, i.e.\ $\sigma^{\ell} (x) = \mathrm{max} (0,x)$, we say that $\mathcal{N}$ is a ReLU FCN.  When $\sigma^{\ell}$ is a step function for each $\ell \in \{1, \ldots, L\}$, i.e.\ $\sigma^{\ell} (x) = 1$ if $x \geq z$ and $0$ otherwise for some $z \in \mathbb{R}$, we say that $\mathcal{N}$ is a step-function FCN. A FCN $\mathcal{N}$ with domain $\Delta$
is \emph{monotonic} is it satisfies the following property:
for all $\mathbf{x}, \mathbf{x}' \in \Delta$, if $x_i \leq x_i'$ for all $i \in \left\{1,...,n\right\}$, then $\Tilde{\mathcal{N}} (\mathbf{x}) \leq \Tilde{\mathcal{N}}(\mathbf{x}')$. Monotonicity is syntactically ensured  by requiring that the weight matrices in all layers of the network are non-negative and that the activation functions in all layers are monotonic.

\paragraph{Attribution Methods. } Attribution methods \cite{axiomaticattrib,manyshapsundararajan20b,shapleyval} are a family of explanation techniques which, given as input function $f : \mathbb{R}^n \mapsto \mathbb{R}$, a vector $\mathbf{x} \in \mathbb{R}^n$ and a baseline vector $\mathbf{x}' \in \mathbb{R}^n$, assign a numerical score or contribution $C_i^f (\mathbf{x},\mathbf{x}')$ to each component $i \in \left\{1,...,n\right\}$.  Attribution methods fulfil some (or all) of the following axioms for all functions
$\mathbb{R}^n \mapsto \mathbb{R}$ and vectors  $\mathbf{x}, \mathbf{x'} \in \mathbb{R}^n$, components $1 \leq i \leq n$ and coefficients $\lambda_1, \lambda_2 \in \mathbb{R}$: \emph{(i)}
    \emph{Completeness:} $f(\mathbf{x}) - f(\mathbf{x}') = \sum_{j=1}^n C_j^f (\mathbf{x}, \mathbf{x}')$; 
    \emph{(ii)} \emph{Zero-contribution:} $C_i^f (\mathbf{x}, \mathbf{x}') = 0$  whenever $f(\mathbf{y}) = f(y_1, ..., y_{i-1}, z, y_{i+1}, ... y_n) $
     for each $\mathbf{y}\in \mathbb{R}^n$ and each $z \in \mathbb{R}$; \emph{(iii)}
    \emph{Symmetry:} $ C_i^f (\mathbf{x}, \mathbf{x}') = C_j^f (\mathbf{x}, \mathbf{x}')$ if $x_i = x_j$, $x_i' = x_j'$ and
     $f(y_1, ..., y_{i}, ..., y_{j}, ... y_n) = f(y_1, ..., y_{j}, ..., y_{i}, ... y_n)$ for each $\mathbf{y}\in \mathbb{R}^n$; and \emph{(iv)} \emph{Linearity:} $C_i^{\lambda_1 f_1 + \lambda_2 f_2} (\mathbf{x}, \mathbf{x}') = \lambda_1 C_i^{f_1} (\mathbf{x},\mathbf{x}') + \lambda_2 C_i^{f_2} (\mathbf{x}, \mathbf{x}')$.
     
Completeness ensures that contributions add up to the change in value of the function. Zero-contribution ensures that 
arguments not influencing the value of the function are assigned $0$ as contribution. Symmetry ensures that arguments playing a symmetric role are assigned the same contribution. Finally, linearity ensures that contributions for
a function expressed as a linear combination of other functions can be computed as a linear combination of their contributions.

A wide range of attribution-based methods has been proposed.
The \emph{Shapley values} method \cite{shapleyval} is one of the most popular thanks to its nice properties. Calculating Shapley values is, however,  intractable, which has motivated research on approximations  \cite{approxshap}. Other popular attribution methods have been designed for neural networks; these include \emph{Layer-wise Relevance Propagation} \cite{layerwiserel},
\emph{DeepLIFT} \cite{deeplift}, \emph{Deep Taylor decompositions} \cite{deeptaylor}, and \emph{Saliency Maps} \cite{saliencymap,sanitysaliencymaps,realtimesaliency,saliencyvariotionaldropout}. 

We will exploit the properties of 
\textit{Integrated Gradients} \cite{axiomaticattrib,aummanshap}, which is 
applicable to continuous functions differentiable almost everywhere. The contribution of
argument $i$ of function $f$ for input vector $\mathbf{x}$ and baseline$\mathbf{x}'$ is defined as follows:
\begin{equation}\label{contribdef}
    C_i^{f} (\mathbf{x}, \mathbf{x}') := (x_i - x'_i) \int_0^1 (\nabla f)_i\left( \mathbf{x}' + \tau(\mathbf{x} - \mathbf{x}') \right) \mathrm{d}\tau.
\end{equation} 
Integrated gradients is the only path-based attribution method satisfying all of the aforementioned axioms \cite{pathadditivecost}. Furthermore, it is well-suited for functions realised by neural networks, which  typically satisfy its continuity
and differentiability requirements.

\paragraph{Explainability queries and explanations.} An explainability query is a Boolean question, formalised as a decision problem,
that we ask about a model and an input vector.

Consider a domain $\Delta \subseteq \mathbb{R}^n$. Given vector $\mathbf{x} \in \Delta$ and  function $f : \Delta \mapsto \{0,1\}$, a
\emph{contrastive explanation} is a subset $S \subseteq \{1,\ldots,n\}$
such that $f(\mathbf{x}^{S|\mathbf{y}})\neq f (\mathbf{x})$ for some 
vector $\mathbf{y} \in \Delta$.
The \emph{Minimum Change Required (MCR)} query
is to decide whether there exists a contrastive explanation  for the given $f$ and $\mathbf{x}$
of size at most a given number $1 \leq k \leq n$.

Given $\mathbf{e} \in \Delta$ and  
$f : \Delta \mapsto \{0,1\}$, an \emph{abductive explanation} is a subset $S \subseteq \{1,\ldots,n\}$
such that $f(\mathbf{e}^{\overline{S}|\mathbf{z}}) = f(\mathbf{e})$ for all $\mathbf{z} \in \Delta$, where $\overline{S} = \{1,\ldots,n\} \setminus S$.
The \emph{Minimum Sufficient Reason (MSR)} query
is to decide whether there exists an abductive explanation for the given function $f$ and vector $\mathbf{e}$
of size at most a given number $1 \leq k \leq n$.

A contrastive (respectively, abductive) explanation  is
cardinality-minimal if there exists no contrastive
(respectively, abductive) explanation of smaller cardinality for the same function and input vector.

\section{Intractability for Monotonic Networks}\label{sec:intractability}

MCR is known to be NP-hard already for FCNs implementing a Boolean function and equipped with ReLU
activations only and a threshold-based classification function
\cite{DBLP:conf/nips/BarceloM0S20}.
In turn, MSR is $\Sigma_2^P$-hard for the same setting.
A natural way to recover tractability   is
to focus on models implementing functions with
specific properties, where monotonicity is a natural requirement in many applications.

It was shown in \cite{DBLP:conf/icml/0001GCIN21}
that subset-minimal abductive explanations for any
ML classifier implementing a monotonic function can be 
computed in polynomial time. This tractability result
is encouraging as well as rather general: it makes no assumptions on the type and structure of the monotonic classifier, or on the domain (e.g., real-valued or Boolean) of the corresponding monotonic function. 

The complexity of MCR and MSR in the monotonic setting, however, remains unclear. On the one hand, the algorithms in \cite{DBLP:conf/icml/0001GCIN21} cannot be used to compute cardinality-minimal 
explanations and thus their tractability results do not imply tractability of MCR and MSR. On the other hand, the neural networks used in the hardness proofs 
in \cite{DBLP:conf/nips/BarceloM0S20} are non-monotonic.

In this section we close this gap and show that, 
surprisingly, both MCR and MSR remain intractable in 
general for monotonic neural networks, already in the Boolean case.

\begin{theorem}\label{th:intractability-mcr-pie-monotonic}
MCR is NP-complete for monotonic Boolean functions implemented by FCNs.
\end{theorem}

\begin{proof}
We show hardness by reduction from SET-COVER, which is the problem of checking, given as input $m$ subsets $E_1, ..., E_m$ of $\left\{1,...,n\right\}$
 such that $ \bigcup_{i \in \left\{1,...,m\right\}} E_i = \left\{1,...,n\right\}$ whether there exists $S \subseteq \{1, \ldots, m\}$ of size at most $K$ such that
$ \bigcup_{i \in S} E_i = \left\{1,...,n\right\}$.

We map an instance $E_1, ..., E_m, K$ of SET-COVER to an instance of MCR for monotonic Boolean functions implemented by FCNs
by setting
$k = K$,  $\mathbf{x} = \mathbf{0}_{m}$, and $\mathcal{N}$ the 2-layer monotonic step-function FCN defined as given next.

In the first layer, 
$\mathbf{W}^1$ is a $(n \times m)$ matrix with values $W_{j,i} =1$ if $j \in E_i$ and 0 otherwise, $\mathbf{b}^1 = \mathbf{0}_n$ and the activation function $\sigma^1$ is a step function $\sigma^1 (z) = 1$ if $z >0$ and $0$ otherwise. In the second layer, $\mathbf{W}^2 = \mathbf{1}_n$, $\mathbf{b}^2 = 0$, $\sigma^2$ is the step function $\sigma^2 (z) = 1$ if $z \geq n$ and $0$ otherwise, and  $\mathrm{cls}_{\{0,1\}}$ is the identity. One can verify that $\mathcal{N}(\mathbf{x}) = 0$.

Assume there exists
$S \subseteq \left\{1,...,m\right\}$  of cardinality at most $k = K$ and  $\mathbf{y} \in \{0,1\}^m$ satisfying $\mathcal{N}(\mathbf{x}^{S|\mathbf{y}}) \neq \mathcal{N}(\mathbf{x})$. 
 We claim that, by construction of $\mathcal{N}$, $S$ is a solution to the corresponding SET-COVER instance. Given that $\mathcal{N}$ can only take values $0$ and $1$, we must have $\mathcal{N}(\mathbf{x}^{S|\mathbf{y}}) = 1$, thus $h^2 \geq n$. This enforces that $h^1_j > 0$ for all $j \in \left\{1,...,n\right\}$. By construction, for all $j \in \left\{1,...,n\right\}$, there exists $i \in S$ such that $j \in E_i$, and $y_i > 0$. In particular, this gives us $\bigcup_{i \in S} E_i = \left\{1,...,n\right\}$. 
 
 For the converse, let $S$ be a solution to SET-COVER. We claim that $S$ and
 vector $\mathbf{y} = \overline{\mathbf{0}}_m$ constitute a certificate for 
 the constructed MCR instance. Indeed, $\bigcup_{i \in S} E_i = \left\{1,...,n\right\}$ thus for all $j \in \left\{1,...,n\right\}$, there exists $i \in S$ such that $j \in E_i$. Thus, with input $\mathbf{x}^{S|\overline{\mathbf{0}}_m}$, $h^1_j >0$ for all  $j \in \left\{1,...,n\right\}$. This yields $h^2 =n$ and hence $\mathcal{N}(\mathbf{x}^{S|\mathbf{y}}) = 1 \neq \mathcal{N}(\mathbf{x})$.
 
Membership in NP follows since $S\subseteq \{1, \ldots, m\}$ of cardinality at most $k$ and $\mathbf{y} \in \{0,1\}^m$ provide a certificate: $(S,\mathbf{y})$ witnesses a solution of MCR for input $\mathcal{N}$, $\mathbf{x}$ and $k$ if $\mathcal{N}(\mathbf{x}^{S|\mathbf{y}})  \neq \mathcal{N}(\mathbf{x})$, which is polytime verifiable.
 \end{proof}

\begin{theorem}\label{th:intractability-msr-monotonic}
MSR is NP-complete for monotonic Boolean functions implemented by FCNs.
\end{theorem}

\begin{proof}

    We again show NP-hardness by reduction from SET-COVER. We map an instance $E_1, ..., E_m, K$ of SET-COVER to an instance of MSR 
    by setting $k = K$,  $\mathbf{e} = \overline{\mathbf{0}}_m$, and $\mathcal{N}$ the 2-layer monotonic step-function FCN described in the proof of Theorem \ref{th:intractability-mcr-pie-monotonic}. One can verify that $\mathcal{N}(\mathbf{e}) = 1$.

    Assume there exists $S \subseteq \left\{1,...,m\right\}$  of cardinality at most $k = K$ satisfying $\mathcal{N}(\mathbf{e}^{\overline{S}|\mathbf{z}}) = \mathcal{N}(\mathbf{e})$ for all $\mathbf{z} \in \{0,1\}^m$. We claim that $S$ is a solution to the corresponding SET-COVER instance. In particular, we have $\mathcal{N}(\mathbf{e}^{\overline{S}|\mathbf{0}_m}) = 1$, thus $h^2 \geq n$. This enforces $h^1_j > 0$ for all $j \in \left\{1,...,n\right\}$. By construction, for all $j \in \left\{1,...,n\right\}$, there exists $i \in S$ such that $j \in E_i$, which is equivalent to $\bigcup_{i \in S} E_i = \left\{1,...,n\right\}$.

    For the converse, let $S$ be a solution to SET-COVER. We claim that $S$ is a certificate for the constructed MSR instance. Indeed, $\bigcup_{i \in S} E_i = \left\{1,...,n\right\}$ thus for all $j \in \left\{1,...,n\right\}$, there exists $i \in S$ such that $j \in E_i$. Thus, with input $\mathbf{e}^{\overline{S}|\mathbf{0_m}}$, $h^1_j >0$ for all  $j \in \left\{1,...,n\right\}$. This yields $h^2 =n$ and hence $\mathcal{N}(\mathbf{e}^{\overline{S}|\mathbf{0_m}}) = 1 = \mathcal{N}(\mathbf{e})$. By monotonicity, we thus have $\mathcal{N}(\mathbf{e}^{\overline{S}|\mathbf{z}}) \geq \mathcal{N}(\mathbf{e}^{\overline{S}|\mathbf{0_m}}) = 1$ for all $\mathbf{z} \in \{0,1\}^m$. 
    
    Membership in NP follows since a set $S\subseteq \{1, \ldots, m\}$ of cardinality at most $k$ provides a certificate. Indeed,  MSR is true if $\mathcal{N}(\mathbf{e}^{\overline{S}|\mathbf{0_m}})  = \mathcal{N}(\mathbf{e}) = 1$, or $\mathcal{N}(\mathbf{e}^{\overline{S}|\overline{\mathbf{0}}_m})  = \mathcal{N}(\mathbf{e}) = 0$, which is verifiable in polynomial time.
\end{proof}

Note that, although MSR remains intractable, monotonicity does bring its complexity  down from the second level of the polynomial hierarchy to NP for Boolean functions.

\section{Achieving Tractability of MCR and MSR}

The hardness proofs in Section \ref{sec:intractability} rely on the use of the step activation function which, in contrast to the
activations used in practice such as ReLU, is a discontinuous function.

In this section, we show that cardinality-minimal abductive and contrastive explanations become polytime computable (and hence MSR and MCR become tractable) if we additionally assume that the activations in the FCNs implementing the monotonic function of interest are continuous everywhere and differentiable almost everywhere; these are mild restrictions that are satisfied by most practical activation functions. 

\begin{definition}\label{def:monotonicity-continuity}
An activation function is \emph{admissible} if 
it is continuous everywhere, differentiable almost everywhere, and non-decreasing.
\end{definition}

Furthermore, our tractability results can be extended beyond the Boolean setting to real-valued functions over a bounded domain as defined next.

\begin{definition}
A domain $\Delta \subset \mathbb{R}^n$ is \emph{bounded} if there exist  lower and upper bound vectors
$\mathbf{l} \in \Delta$ and $\mathbf{u} \in \Delta$  such that, for all $\mathbf{x} \in \Delta$ and each $i \in \{1,...,n\}$,  we have $l_i \leq x_i \leq u_i$.
\end{definition}

Note that Boolean domains are bounded by the null vector of the relevant dimension and its complement.

\subsection{Properties of Monotonic Networks with Admissible Activation}

In the remainder of this section, we focus on FCNs where all activations are admissible and where monotonicity is ensured syntactically by requiring that weight matrices in all layers contain only non-negative weights.
Let us therefore fix an arbitrary FCN $\mathcal{N}$ over a domain $\Delta \subseteq \mathbb{R}^n$ of dimension $n$ satisfying these requirements, which we exploit in the formulation of our 
results; furthermore, assume that $\Delta$
is bounded by a lower bound vector $\mathbf{l}$ and upper bound vector $\mathbf{u}$. 

The continuity and differentiability requirements of the activation functions ensure that the gradient of 
$\Tilde{\mathcal{N}}$ can be computed for each input feature vector $\mathbf{x}$. 
In turn, as we show next, the monotonicity requirement ensures that each component of the gradient of $\Tilde{\mathcal{N}}$ (the output of the last layer) at $\mathbf{x}$ can be expressed as a sum where  \emph{(1)} the number of elements in the sum is fixed for $\mathcal{N}$ (i.e., it does not depend on $\mathbf{x}$) and it is the same for all vector components; and \emph{(2)} 
each element of the sum consists of a product involving a value that depends on $\mathbf{x}$ but which is \emph{always non-negative}, and two coefficients that do not depend on $\mathbf{x}$.
This key property of the gradient allows us to exploit the theoretical properties of the integrated gradients attribution method.
In particular, we can show that, by setting a component $x_i$ of the input vector $\mathbf{x}$ to the corresponding component of the lower or upper bound vector, depending on whether the prediction for $\mathbf{x}$ is $1$ or $0$, we are not altering the relative order of the 
integrated gradient attributions for the remaining components. 

These properties, which are established by the following technical lemma, 
constitute the basis of our greedy algorithms for answering explainability queries.


\begin{lemma}\label{lem:auxiliary}
There exists an integer $M \in \mathbb{Z}_{\geq 0}$, positive coefficients $\{A_m\}_{1 \leq m\leq M}$, and $\{B_{i}\}_{1 \leq i \leq n}$, and functions $\{g_m\}_{1 \leq m \leq M}$ from $\Delta$ to $\mathbb{R}_{\geq 0}$, such that the following identities are satisfied for each
$\mathbf{x}, \mathbf{x}' \in \Delta$ and each $i,j \in \left\{1,...,n\right\}$
\begin{equation}
 (\nabla \Tilde{\mathcal{N}})_i (\mathbf{x})   =  B_i \; \sum_{m=1}^M A_m \;  g_m (\mathbf{x})\label{eq:nabla}
 \end{equation}
and 
 \begin{multline}
 \Tilde{\mathcal{N}}(\mathbf{x}^{\{i\}|\mathbf{x}'}) - \Tilde{\mathcal{N}}(\mathbf{x}^{\{j\}|\mathbf{x}'})   = \\  (B_i (x_i' - x_i) - B_j (x_j' - x_j) \sum_{m=1}^M \! A_m \! \int_0^1 \!  g_m (\mathbf{p}^{ij} (\tau)) \mathrm{d}\tau, \label{eq:difference}
\end{multline}
where $\mathbf{p}^{ij} (\tau) = \mathbf{x}^{\left\{j\right\}|\mathbf{x}'} + \tau(\mathbf{x}^{\left\{i\right\}|\mathbf{x}'} - \mathbf{x}^{\left\{j\right\}|\mathbf{x}'})$.
\end{lemma}

\begin{proof}
We show \eqref{eq:nabla} by induction on the number of layers $L$ in $\Tilde{\mathcal{N}}$.
If $L=1$, then $\Tilde{\mathcal{N}} = \langle \mathbf{W}, b, \sigma \rangle$ with $\mathbf{W} \in \mathbb{R}^n$, $b \in \mathbb{R}$ and
$\sigma: \mathbb{R} \mapsto \mathbb{R}$. By the chain rule, 
$(\nabla \Tilde{\mathcal{N}})_i (\mathbf{x}) = W_i \cdot (D\sigma)(\mathbf{W} \cdot \mathbf{x} + b)$,
with $D\sigma$ the derivative of $\sigma$ in Euler's notation.
This is of the form \eqref{eq:nabla} with $M=1$, 
$A_1 = 1 \geq 0$, $B_i = W_i$, and $g_1(\mathbf{x}) = (D\sigma)(\mathbf{W} \cdot \mathbf{x} + b)$. Monotonicity of $\sigma$ ensures $g_1(\mathbf{x}) \geq 0$ for any $\mathbf{x}$.

For the inductive case, assume \eqref{eq:nabla} holds for each network with
$L-1$ layers satisfying the same requirements as $\mathcal{N}$. The application of  $\Tilde{\mathcal{N}} = \langle \{\mathbf{W}^{\ell}\}_{1 \leq \ell \leq L}, \{\mathbf{b}^{\ell}\}_{1 \leq \ell \leq L}, \{\sigma^{\ell}\}_{1 \leq \ell \leq L\}}  \rangle$ with $L$ layers to  $\mathbf{x}$ is defined as
$\sigma^L(h^L(\mathbf{x}))$. By the
chain rule and the definition of
$h^L$ we obtain the following identity:
\begin{multline}\label{eq:layer-L}
            (\nabla \Tilde{\mathcal{N}})_i (\mathbf{x}) = (\nabla h^L)_i(\mathbf{x}) \cdot (D \sigma^L)(h^{L}(\mathbf{x}))   = \\ \sum_{j = 1}^{d_{L-1}} W_{j}^L \cdot (\nabla \Tilde{\mathcal{N}}^j)_i(\mathbf{x}) \cdot (D\sigma^L)(h^{L}(\mathbf{x})).
    \end{multline}
Here, $\Tilde{\mathcal{N}}^j$ is given by
weight matrices $\{\mathbf{W}^{\ell}\}_{1 \leq \ell \leq L-2}$ and $\mathbf{W}_j^{L-1}$ (representing the 
$j$-th row of $\mathbf{W}^{L-1}$), 
bias vectors
$\{\mathbf{b}^{\ell}\}_{1 \leq \ell \leq L-2}$ and $b_j^{L-1}$ (representing the $j$-th element of $\mathbf{b}^{L-1}$), and activations
$\{\sigma^{\ell}\}_{1 \leq \ell \leq L-1\}}$.
We apply the inductive hypothesis to compute the gradient
for each $1 \leq j \leq d_{L}-1$, which is  
$(\nabla \Tilde{\mathcal{N}}^j)_i(\mathbf{x}) = B_i^j \; \sum_{m_{j}=1}^{M_j} A_{m_j}^j \;  g_{m_j}^j(\mathbf{x})$.
But now, we can replace the value of the gradients in the sum of  \eqref{eq:layer-L} with these values and show the statement of the lemma by instantiating  \eqref{eq:nabla} with
$M = \sum_{j = 1}^{d_{L-1}} M_j$, $A_m = W^L_j A_{m_j}^j$, $B_i = \; B^j_i$ and $g_{m}(\mathbf{x}) = \; g_{m_j}^j (\mathbf{x}) \cdot (D \sigma^L)(h^{L}(\mathbf{x}) ) $. Again, by induction,  $A_m \geq 0$ and $g_m(\mathbf{x}) \geq 0$ for each $m$ and $\mathbf{x}$.

We now show \eqref{eq:difference}. Let us consider  the attribution for $\Tilde{\mathcal{N}}$ defined in \eqref{contribdef}. Assume $i \neq j$ (otherwise the equation holds trivially). By replacing the gradient in  \eqref{contribdef} with \eqref{eq:nabla}, the value of 
    $C_i^{\Tilde{\mathcal{N}}} (\mathbf{x}, \mathbf{x}')$ is  $B_i (x_i - x'_i) \; \sum_{m=1}^M A_m \; \int_0^1 g_m(\mathbf{x}' + \tau(\mathbf{x} - \mathbf{x}')) \mathrm{d}\tau.$ 
Since integrated gradients satisfy the completeness and zero contribution axioms, we can compute the difference $\Tilde{\mathcal{N}}(\mathbf{x}^{\left\{i\right\}|\mathbf{x}'}) - \Tilde{\mathcal{N}}(\mathbf{x}^{\left\{j\right\}|\mathbf{x}'})$ as the sum of contributions $C_i^{\Tilde{\mathcal{N}}} (\mathbf{x}^{\left\{i\right\}|\mathbf{x}'}, \mathbf{x}^{\left\{j\right\}|\mathbf{x}'})$ and $C_j^{\Tilde{\mathcal{N}}} (\mathbf{x}^{\left\{i\right\}|\mathbf{x}'}, \mathbf{x}^{\left\{j\right\}|\mathbf{x}'})$
to obtain 
\begin{multline}\label{eq:int-expression}
(x'_i - x_i) \int_0^1 (\nabla \Tilde{\mathcal{N}})_i\left( \mathbf{p}^{ij} (\tau) \right) \mathrm{d}\tau -  \\
(x'_j - x_j) \int_0^1 (\nabla \Tilde{\mathcal{N}})_j\left( \mathbf{p}^{ij} (\tau) \right) \mathrm{d}\tau.
\end{multline}
%
The gradients $(\nabla \Tilde{\mathcal{N}})_i (\mathbf{p}^{ij} (\tau))$ and
$(\nabla \Tilde{\mathcal{N}})_j (\mathbf{p}^{ij} (\tau))$ are provided by \eqref{eq:nabla}; when replaced in
\eqref{eq:int-expression}, they  yield \eqref{eq:difference}.
%
\end{proof}

\subsection{Algorithms for Computing Explanations}\label{sec:tractability}

We are now ready to present our greedy algorithms for computing cardinality-minimal explanations in polynomial time.

Algorithm \ref{greedy algo mcr} takes as input a monotonic FCN $\mathcal{N}$ with admissible activation functions over a bounded domain $\Delta$ with lower bound $\mathbf{l}$ and upper bound $\mathbf{u}$, and an input feature vector $\mathbf{x} \in \Delta$, and computes a cardinality-minimal contrastive explanation as detailed next. 

The algorithm first applies the model $\mathcal{N}$ to the input vector $\mathbf{x}$ and, based on the obtained prediction, chooses to consider the domain's lower bound vector $\mathbf{l}$ (if $\mathcal{N}(\mathbf{x}) = 1$) or the upper bound vector $\mathbf{u}$ (if $\mathcal{N}(\mathbf{x}) = 0$) when searching for vectors that change the model's prediction. The monotonicity requirement will ensure that no other vectors need to be considered. 
Then, in each iteration of the first loop, the algorithm sets each individual input feature to the value of the relevant bound vector (while leaving the remaining components unchanged) and applies the input model to the resulting vector.
 The values obtained by each of these applications of the model are then sorted in ascending or descending order depending on the value of
 $\mathcal{N}(\mathbf{x})$.
In the second loop, the algorithm successively assigns the components of $\mathbf{x}$ to the chosen bound vector in the order established in the previous step until the prediction changes. The algorithm then returns $S$ 
 consisting of all
 features that were set to the chosen bound.

 Our algorithm is quadratic in the number of input features: both loops require linearly many
applications of the FCN, and each application is feasible in linear time in the number of features \cite{deeplearningbook}. The algorithm's correctness relies on  \eqref{eq:difference} in Lemma \ref{lem:auxiliary}, which ensures that, when set to the chosen bound, each of the features selected by the algorithm in the second loop yields the largest change (amongst all other possible feature choices) in the application of the model, thus getting  as close as possible to the prediction threshold. As a result, the output 
subset $S$ is guaranteed to contain a smallest number of features.

\begin{algorithm}[tb]
    \caption{Computing  contrastive explanations.}
    \label{greedy algo mcr}
    \textbf{Input:} vector $\mathbf{x} \in \Delta$ with $\Delta$  bounded by vectors $\mathbf{l}, \mathbf{u}$,  and monotonic FCN $\mathcal{N}: \Delta \mapsto \{0,1\}$ with admissible activation functions. \\
    \textbf{Output:} A cardinality-minimal contrastive explanation $S$ for $\mathcal{N}$ and $\mathbf{x}$
    \begin{algorithmic}[1]
        \IF{$\mathcal{N}(\mathbf{x}) = 1$}
            \STATE $\mathbf{x}' \gets \mathbf{l}$
        \ELSE
            \STATE $\mathbf{x}' \gets \mathbf{u}$
        \ENDIF
        \FOR{$ 1 \leq j \leq n$}
        \STATE $c_j \gets  \Tilde{\mathcal{N}}(\mathbf{x}^{\{j\}\mid \mathbf{x}'})$
        \ENDFOR
        \STATE $I \leftarrow$ list of indices obtained from sorting $\{c_j\}_{1 \leq j \leq n}$ in ascending (respectively, descending) order with ties broken arbitrarily if $\mathcal{N}(\mathbf{x}) = 1$ (respectively, if $\mathcal{N}(\mathbf{x}) = 0$). 
        \STATE $S \gets \emptyset$
        \FOR{$ 1 \leq j \leq n$}
        \STATE $S \leftarrow S \cup I[j]$
        \STATE \textbf{if} $\mathcal{N}(\mathbf{x}^{S \mid \mathbf{x}'}) \neq \mathcal{N}(\mathbf{x})$  \textbf{then}  \textbf{return} $S$
        \ENDFOR
    \end{algorithmic}
\end{algorithm}

\begin{theorem}\label{th:algo mcr}
    Algorithm \ref{greedy algo mcr} computes 
    a cardinality-minimal contrastive explanation for the input $\mathcal{N}$ and $\mathbf{x}$. 
\end{theorem}

\begin{proof}
By symmetry of the algorithm, we can assume, without loss of generality that $\mathcal{N} (\mathbf{x}) =1$. By monotonicity of $\mathrm{cls}$, it suffices to show that, for each $j \in \left\{1,...,n\right\}$, the choice of $I[j]$ in the second loop yields the largest change in the evaluation of $\Tilde{\mathcal{N}}$ (the output of the last layer). That is, for each $1 \leq j \leq n$ and  $j \leq k \leq n$ we have
$\Tilde{\mathcal{N}} (\mathbf{x}^{S|\mathbf{x}'}) - \Tilde{\mathcal{N}} (\mathbf{x}^{\left(S \cup I[j]\right)|\mathbf{x}'}) \geq \Tilde{\mathcal{N}} (\mathbf{x}^{S|\mathbf{x}'}) - \Tilde{\mathcal{N}} (\mathbf{x}^{\left(S \cup I[k]\right)|\mathbf{x}'})$.
By construction of list $I$, the  inequality $ \Tilde{\mathcal{N}} (\mathbf{x}^{I[j]|\mathbf{x}'}) - \Tilde{\mathcal{N}} (\mathbf{x}^{I[k]|\mathbf{x}'}) \leq 0$ holds for each $1 \leq j \leq k \leq n$. We apply  \eqref{eq:difference} in Lemma \ref{lem:auxiliary} together with the fact that $A_m \geq 0$ and $g_m(\mathbf{x}) \geq 0$ for each $m$ and $\mathbf{x}$ (and hence
$\int_0^1 g_m (\mathbf{p}^{I[j] \; I[k]} (\tau)) \mathrm{d}\tau \geq 0$) to obtain 
$    (B_{I[j]} (x'_{I[j]} - x_{I[j]}) - B_{I[k]} (x'_{I[k]} - x_{I[k]})) \leq 0$.
Since $\left\{I[j], I[k]\right\} \subseteq \overline{S}$, we have $\left(\mathbf{x}^{S|\mathbf{x}'}\right)_{I[j]} = x_{I[j]}$ and $\left(\mathbf{x}^{S|\mathbf{x}'}\right)_{I[k]} = x_{I[k]}$.
By applying \eqref{eq:difference} and the previous inequality,
we finally obtain   $\Tilde{\mathcal{N}} (\mathbf{x}^{\left(S \cup I[j]\right)|\mathbf{x}'}) \leq \Tilde{\mathcal{N}} (\mathbf{x}^{\left(S \cup I[k]\right)|\mathbf{x}'})$. This ensures that the output $S$ is a cardinality-minimal contrastive explanation for $\mathcal{N}$ and $\mathbf{x}$, as required.
%
\end{proof}

Contrastive and abductive explanations are dual to one another. Therefore, as we show next, a minor modification of Algorithm \ref{greedy algo mcr} that exchanges the roles of vectors $\mathbf{x}$ and $\mathbf{x}'$ in the second loop can be used to compute cardinality-minimal abductive explanations.

\begin{theorem}
    A modified version of Algorithm \ref{greedy algo mcr} where $\mathbf{x}$ and $\mathbf{x}'$ are interchanged in Lines 7, 9 and 13 computes 
    a cardinality-minimal abductive explanation for the given input  $\mathcal{N}$ and $\mathbf{x}$. 
\end{theorem}

\begin{proof}
The proof is anlaogous to that of Theorem \ref{th:algo mcr} in showing that the choice of $I[j]$ in the second loop yields the largest change in the evaluation of $\Tilde{\mathcal{N}}$, thus ensuring that $S$ is a cardinality-minimal subset such that $\mathcal{N}(\mathbf{x}'^{S \mid \mathbf{x}}) \neq \mathcal{N}(\mathbf{x}')$. By the choice of $\mathbf{x}'$, we have ensured that $\mathcal{N}(\mathbf{x}') \neq \mathcal{N}(\mathbf{x})$, which gives us that $\mathcal{N}(\mathbf{x}^{\overline{S} \mid \mathbf{x}'}) = \mathcal{N}(\mathbf{x})$. Furthermore, by monotonicity, $\mathcal{N}(\mathbf{x}^{\overline{S} \mid \mathbf{z}}) = \mathcal{N}(\mathbf{x})$ for all $\mathbf{z} \in \Delta$ if and only if $\mathcal{N}(\mathbf{x}^{\overline{S} \mid \mathbf{l}}) = \mathcal{N}(\mathbf{x}) = 1$ or $\mathcal{N}(\mathbf{x}^{\overline{S} \mid \mathbf{u}}) = \mathcal{N}(\mathbf{x}) = 0$. This ensures that $S$ is a minimal abductive explanation.
\end{proof}

Note that, although the correctness of our algorithms relies on the properties of integrated gradients, the algorithms themselves do not compute attribution values, and only rely on the ability to apply the input model as a `black box'.

\section{Discussion and Further Implications}

The notion of cardinality-minimal contrastive explanation 
is closely related to existing notions of \emph{robustness} for ML model predictions proposed in the literature. In particular, the minimality 
requirement ensures that no smaller subset of the features can be used to change the prediction for a given model $\mathcal{N}$ and  input feature vector $\mathbf{x}$; thus, the larger the size of the smallest contrastive explanation, the more robust the prediction of $\mathcal{N}$ on $\mathbf{x}$ is. The notion of D-robustness \cite{DBLP:conf/kr/ShiSDC20} is an instance-based robustness measure based precisely on this idea. The D-ROBUST query can be formalised as the complement of the MCR query: 
given $\mathcal{N}$, $\mathbf{x}$, and $k$, decide whether the size of a cardinality-minimal constrastive explanation for $\mathcal{N}$ and $\mathbf{x}$ is at least $k$.  It was shown in \cite{DBLP:conf/kr/ShiSDC20} that D-ROBUST is coNP-complete for Boolean functions. Thus, Theorem \ref{th:intractability-mcr-pie-monotonic} in Section \ref{sec:intractability} refines the complexity lower bound in \cite{DBLP:conf/kr/ShiSDC20} to monotonic Boolean functions realised by FCNs with step activations. Furthermore, our results in Section \ref{sec:tractability} imply tractability of D-ROBUST for monotonic functions implemented by FCNs with admissible activations and bounded domains. 

Our results also have interesting implications on the problem of 
constructing neural networks that exactly replicate a given function \cite{neuralnetintrac1,neuralnetintrac2,neuralnetintrac3,DBLP:journals/corr/abs-1905-11428}.
In particular, our results imply that, unless P $=$ NP, 
there is no polynomial time algorithm that, given as input a monotonic FCN with step activations, constructs an FCN with admissible activations realising the same function. Indeed, otherwise we could solve in polynomial time the MCR query for monotonic FCNs with step functions by first rewriting the model using admissible activations and then applying our greedy algorithm to the transformed model.

\section{Experiments}

\begin{figure*}[t]
\setlength{\lineskip}{2pt}
\begin{center}
\includegraphics[height=3.0cm]{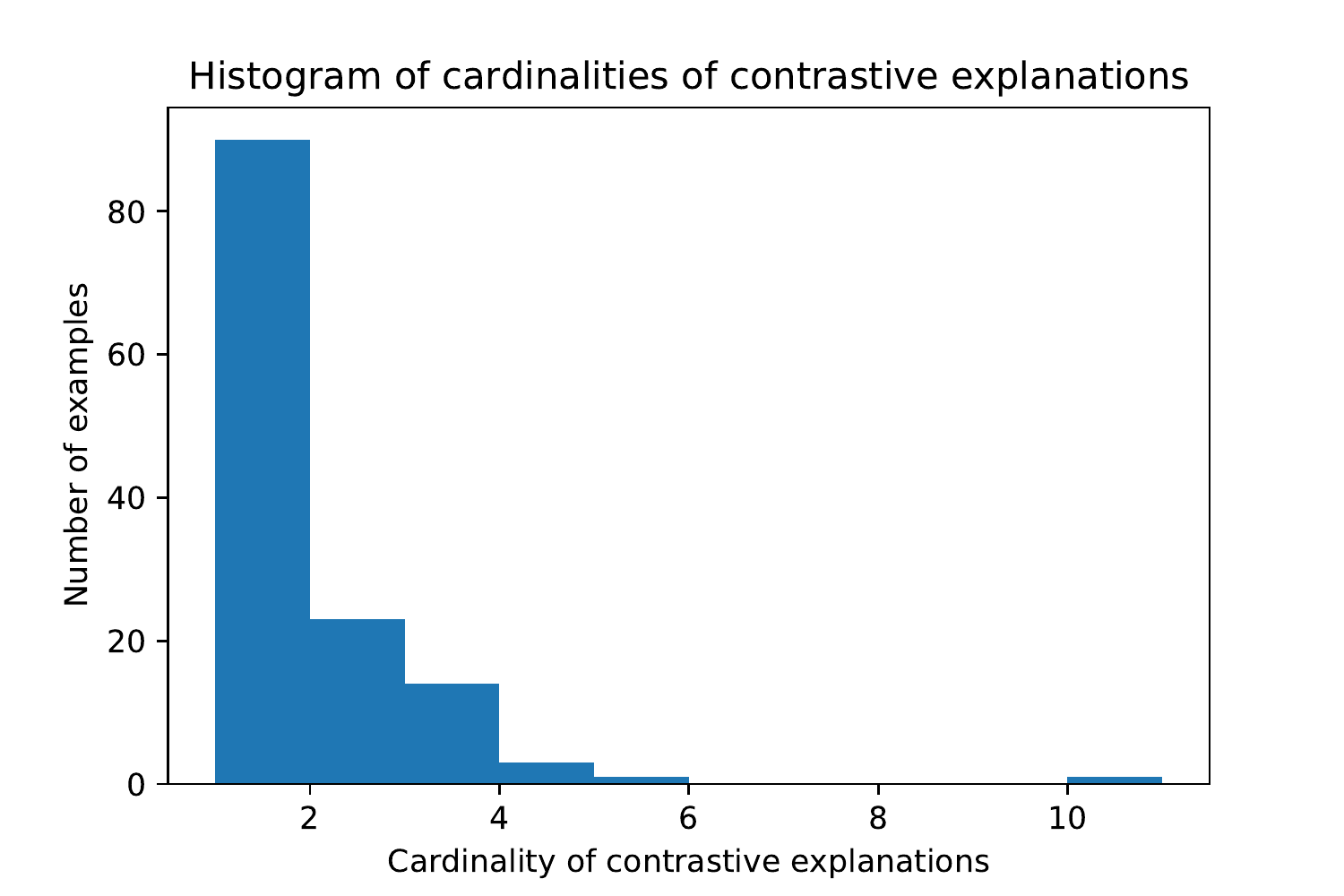}
\includegraphics[height=3.0cm]{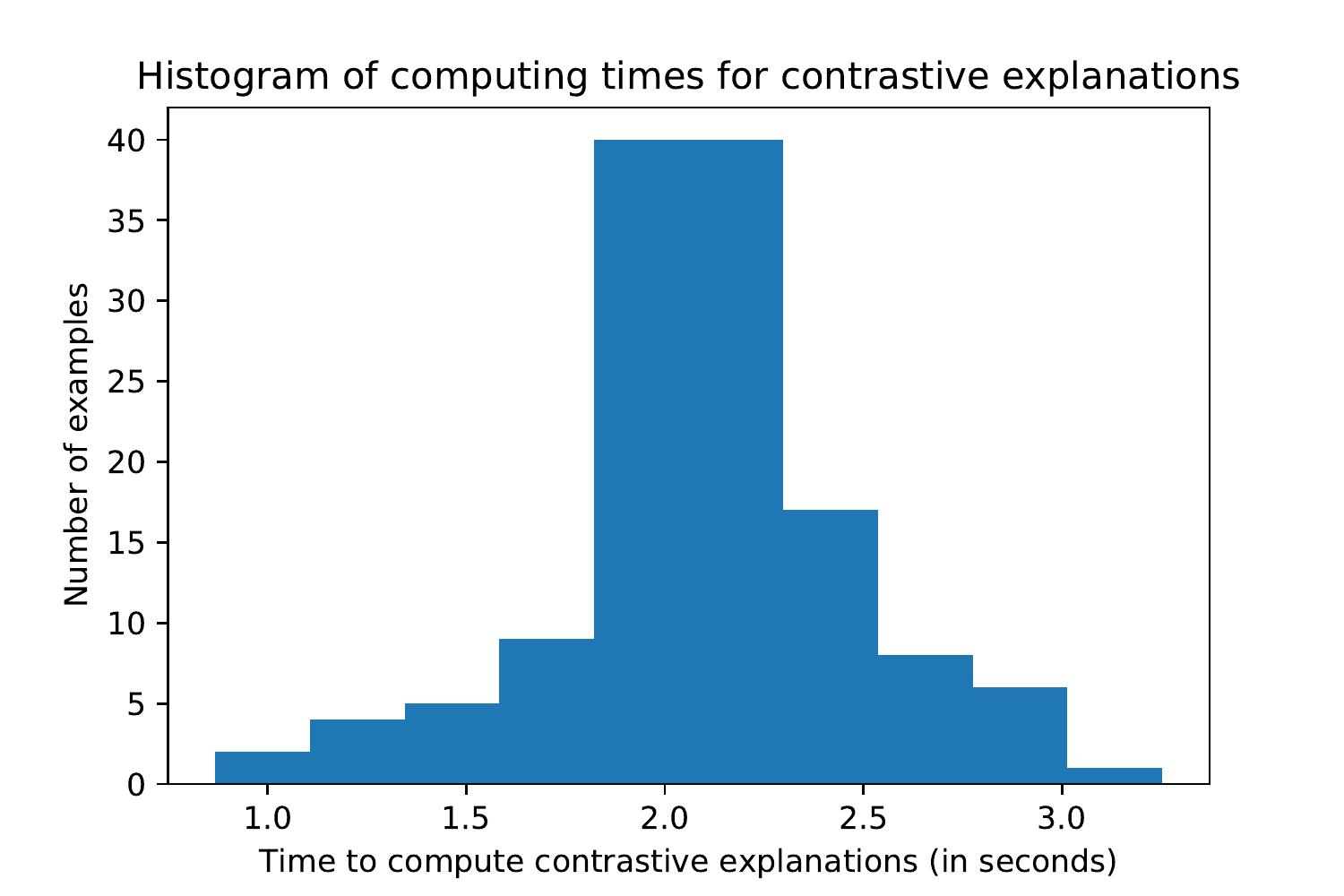}\\
\includegraphics[height=3.0cm]{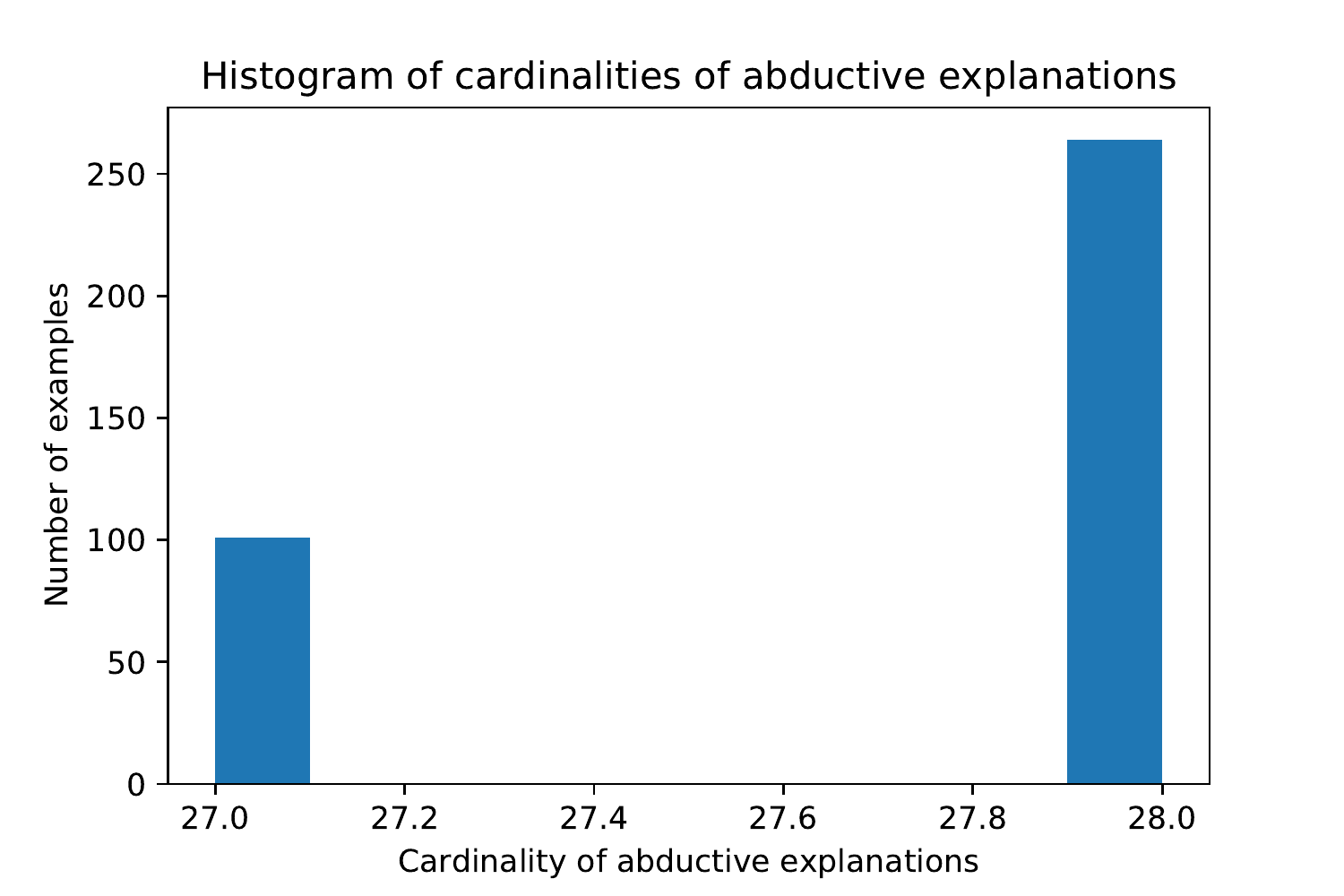}
\includegraphics[height=3.0cm]{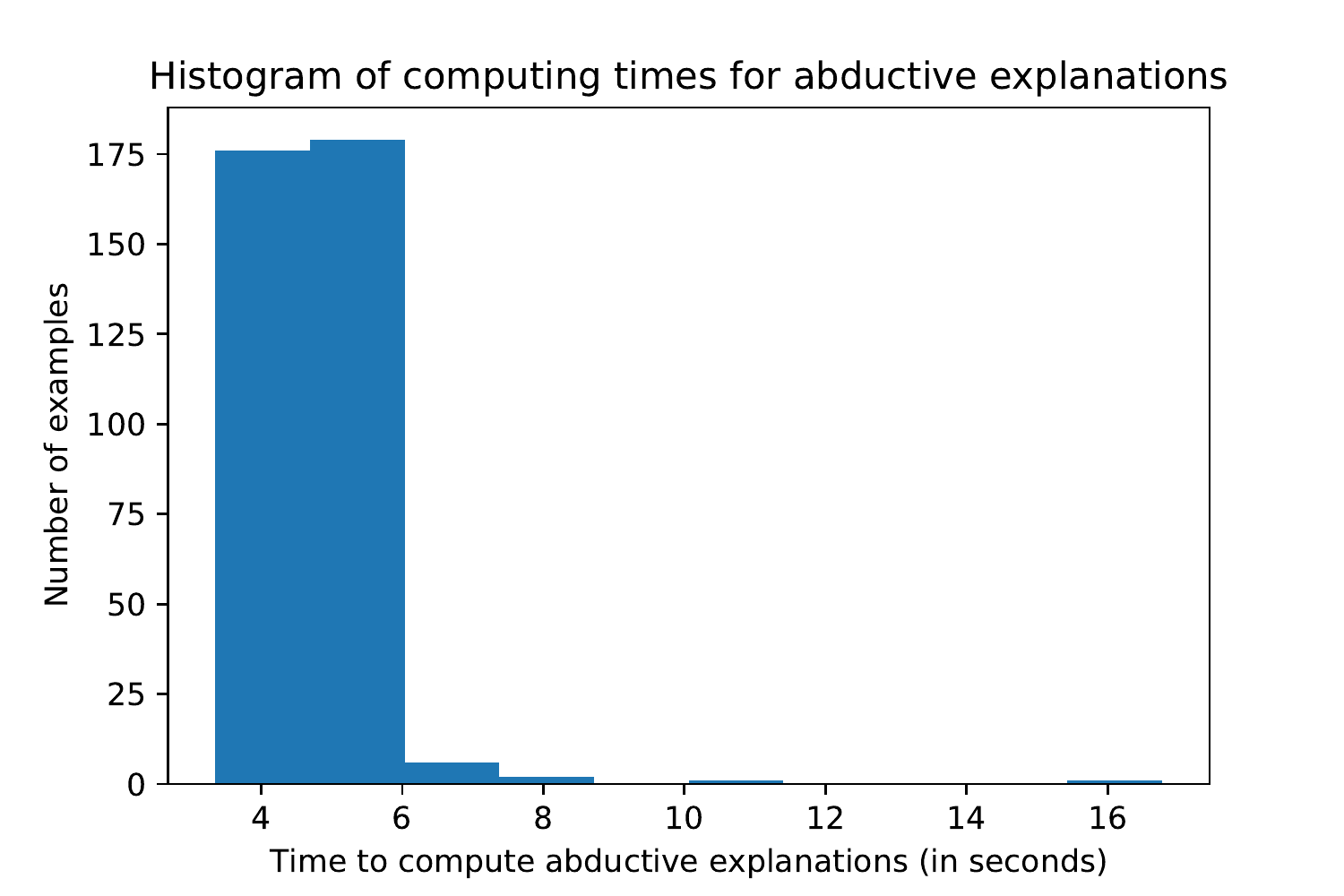}\\
\caption{Explanation sizes and computing times on Loan Defaulter.}
\label{histogramsloan}
\end{center}
\end{figure*}

\begin{figure*}[t]
\setlength{\lineskip}{2pt}
\begin{center}
\includegraphics[height=3.0cm]{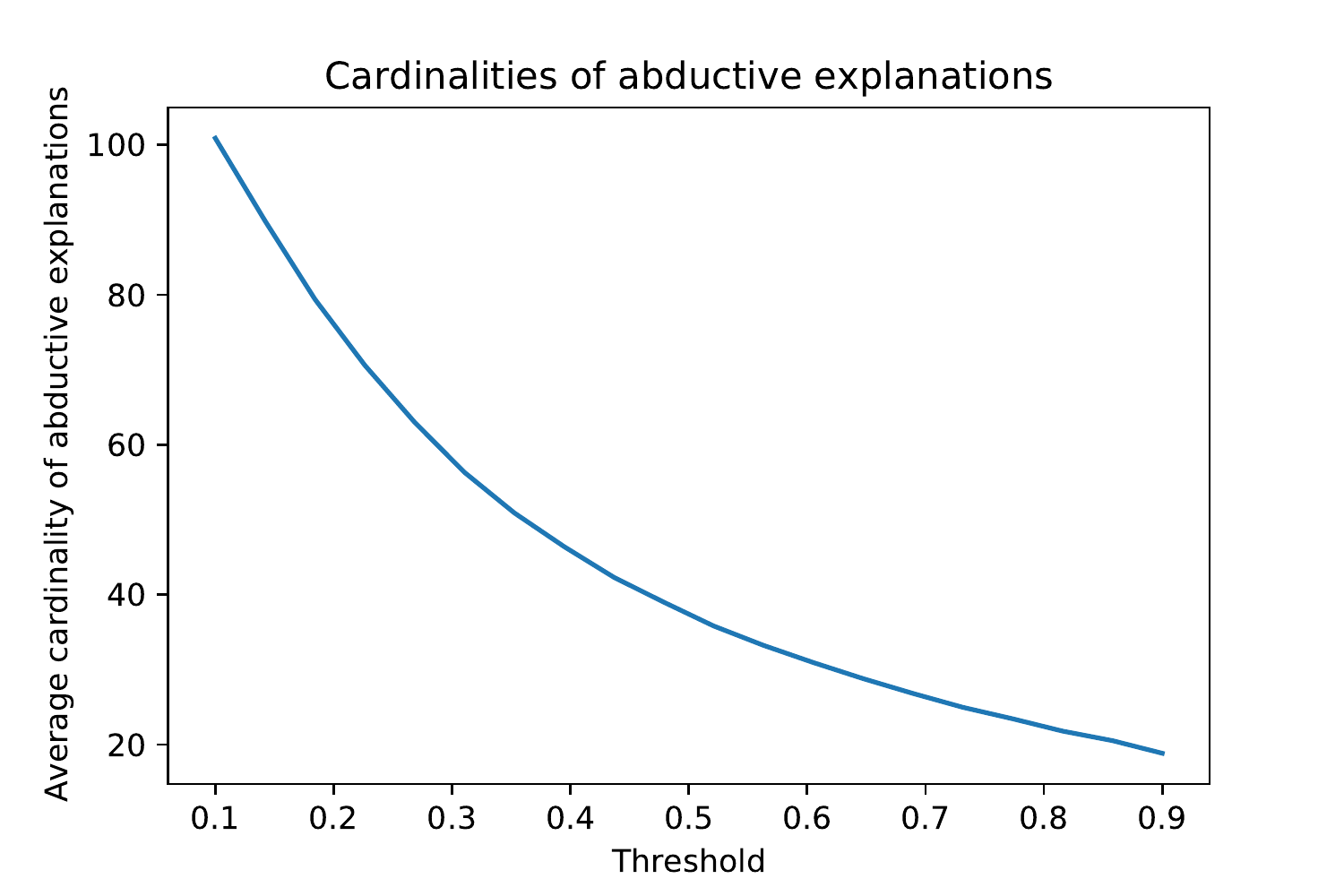}
\includegraphics[height=3.0cm]{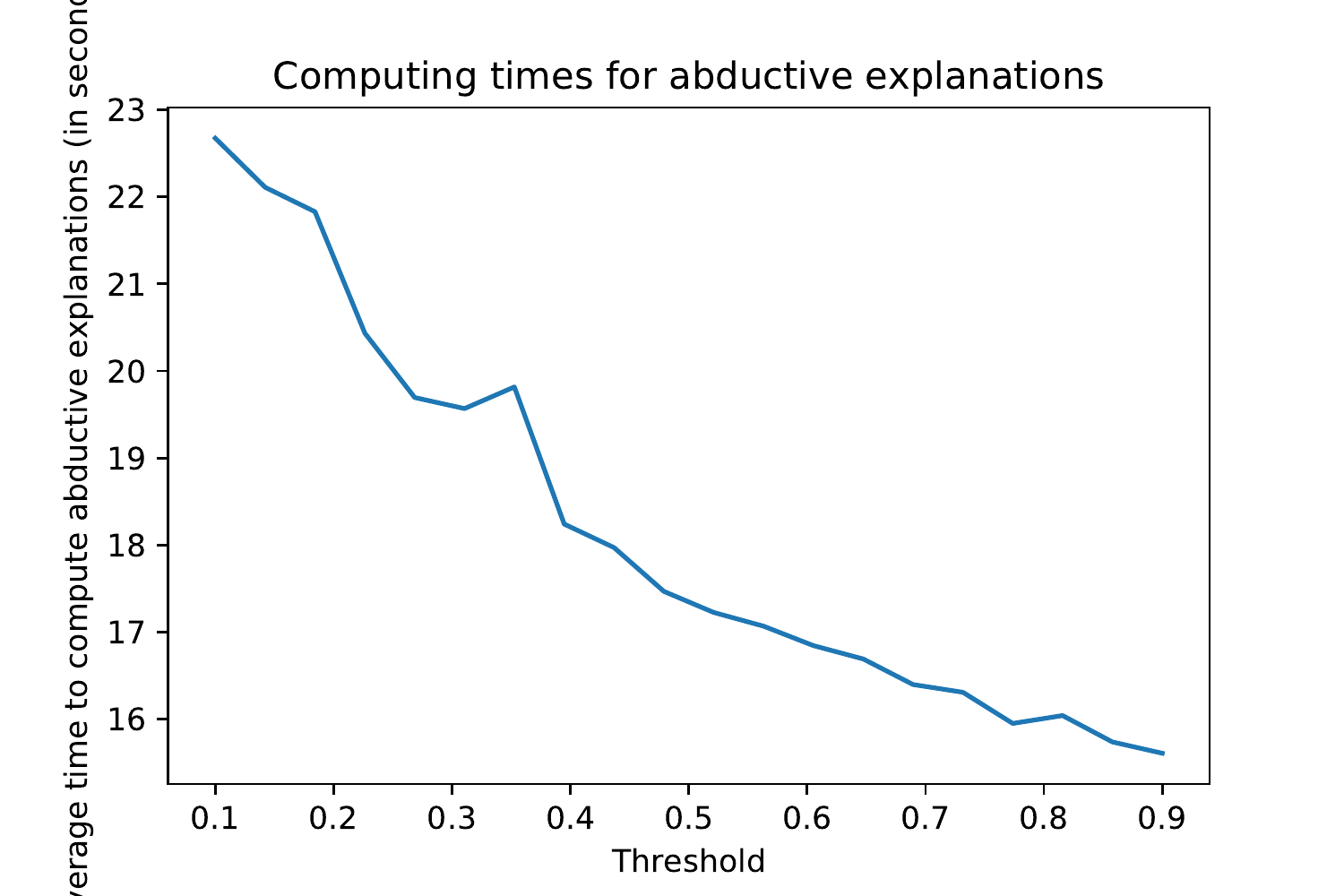}\\
\caption{Explanation sizes and computing times on Blog Feedback.}
\label{plotblog}
\end{center}
\end{figure*}


We have implemented our greedy algorithms for computing cardinality-minimal explanations in Section \ref{sec:tractability}
and assessed their practical suitability on well-known benchmark datasets commonly used to evaluate monotonic models. 
To the best of our knowledge, our implementation is the only one available for computing cardinality-minimal explanations and hence we could not find a suitable benchmark for comparison. 
All experiments were conducted using  Google Colab with GPU.

\paragraph{Datasets.} The \emph{Blog Feedback Regression} dataset \cite{blogreg} is a numeric dataset assembled from $37,279$ blog pages extracted from $1,200$ different sources. The objective is to predict the number of feedbacks that a blog page will receive in a given time window. The $276$ features include the number of links and feedbacks in the past, time and day of publication, discriminative bag of words, etc. 

\emph{Loan Defaulter} is a numeric dataset assembled from LendingClub (a large online loan marketplace). The objective is to predict if the applicant will repay the loan or default. The $28$ features include loans executed in the past, amount of the loan and instalments, applicant's address and zip code, etc.

The features in both datasets are preprocessed and bounded between $0$ and $1$.

\paragraph{Methodology.} We trained monotonic FCN models on both datasets with PyTorch \cite{pytorch} using the mean-squared error loss for the Blog Feedback dataset and the binary cross entropy loss for the Loan Defaulter dataset. We trained the models with Adam \cite{kingma2014adam} for $10$ epochs, setting all negative weights to $0$ after each iteration of Adam to ensure monotonicity. We were able to reach a root mean-squared error (RMSE) of $0.175$ on the test set for the Blog Feedback regression (an acceptable performance given that the state-of-the-art is at $0.158$  \cite{certifiedmonotonicneurips2020}) and reached an accuracy of $60 \%$ on Loan Defaulter (state-of-the-art performance is $65.2\%$). Since the datasets are only partially-monotonic, we should not expect state-of-the-art performance with a fully-monotonic model; please note, however, that the objective of our experiments is not to improve on the state-of-the-art regression and classification metrics, but rather to show that cardinality-minimal explanations for the trained models can be efficiently computed. 

Using the trained monotonic FCNs, we then computed cardinality-minimal abductive and contrastive explanations using the greedy algorithms in Section \ref{sec:tractability}. Note that, although the models for Blog Feedback are trained for regression, our algorithms  
can still seamlessly be applied provided that we introduce a threshold. Indeed, given a FCN $\mathcal{N}: \Delta \mapsto \mathbb{R}$, an input vector $\mathbf{x} \in \Delta$ and a numeric threshold $t$ such that $\mathcal{N} (\mathbf{x}) > t$ (respectively, $\leq t$), Algorithm \ref{greedy algo mcr} can be used to compute a cardinality-minimal subset $S$ such that  $\mathcal{N}(\mathbf{x}^{S \mid \mathbf{x}'}) \leq t$ (respectively, $> t$) and a cardinality-minimal $S$ such that  $\mathcal{N}(\mathbf{x}^{\overline{S} \mid \mathbf{z}}) > t$ (respectively, $\leq t$) for all $\mathbf{z} \in \Delta$.

\paragraph{Results.}

For Loan Defaulter,  contrastive explanations took $1.5$s on average to compute  and contained $1.8$ out of $28$ features on average; in turn, abductive explanations took $5s$ on average  to compute and contained $27.6$ features on average. 
Note that contrastive explanations were much smaller than abductive explanations; this is not unexpected in a partially monotonic dataset since the condition required from abductive explanations requires the prediction to hold \emph{for all possible values} of the features outside the explanation (a very strong condition). Figure \ref{histogramsloan} depicts the cardinalities and computation times for both types of explanations.

For the Blog Feedback Regression dataset, we varied the threshold between the lower bound and the upper bound of the targets and computed the average cardinality and computation time with respect to the threshold for both types of explanations. For contrastive explanations, the cardinality remained constant equal to 1: it was always possible to modify one feature and change the prediction. In the plots of Figure \ref{plotblog}, we focused on abductive explanations and instances $\mathbf{x}$ such that $\mathcal{N} (\mathbf{x}) \leq t$ and, as expected, we can see that the sizes of explanations decrease as the threshold increases.

\section{Conclusion and Future Work}

In this paper, we have studied the problem of computing cardinality-minimal contrastive and abductive explanations for the predictions of  monotonic neural models.
We have strengthened existing intractability results  \cite{DBLP:conf/nips/BarceloM0S20} to the context of monotonic fully-connected networks and proposed additional requirements to regain tractability.  

Our results are of practical interest for the computation of explanations with formal guarantees in the context of monotonic or partially-monotonic tasks. Furthermore,
from a theoretical perspective, our results not only strengthen existing intractability results, but to the best of our knowledge they also provide the first polytime algorithms for computing cardinality-minimal explanations in the context of neural models. Finally, our results also establish a novel connection between the theory of attribution methods and the theory of abductive and contrastive explanation methods, and have interesting implications for other related problems. We hope that our work will motivate further studies on the mathematical properties of explanation methods in Machine Learning.

\section*{Acknowledgments}

This research was supported in whole or in part by
the EPSRC projects OASIS (EP/S032347/1), ConCuR (EP/V050869/1)
and UK FIRES (EP/S019111/1),
the SIRIUS Centre for Scalable Data Access, and
Samsung Research UK.
For the purpose of Open Access, the authors have applied a CC BY public copyright licence to any 
Author Accepted Manuscript (AAM) version arising from this submission.

\bibliographystyle{named}
\bibliography{ijcai23}

\end{document}